\documentclass{article}

\usepackage[preprint]{neurips_2021}

\usepackage[utf8]{inputenc} 
\usepackage[T1]{fontenc}    
\usepackage{hyperref}       
\usepackage{url}            
\usepackage{booktabs}       
\usepackage{amsfonts}       
\usepackage{nicefrac}       
\usepackage{microtype}      
\usepackage{xcolor}         

\usepackage[linesnumbered, ruled]{algorithm2e}
\usepackage{epsfig} 

\usepackage{graphicx}

\usepackage{algpseudocode}
\usepackage{amsmath}
\usepackage{subfig}

\usepackage{amsfonts,amssymb}
\usepackage{multirow}

\usepackage[linesnumbered, ruled]{algorithm2e}

\title{Ranking Cost: Building An Efficient and Scalable Circuit Routing Planner with Evolution-Based Optimization}

\author{%
  Shiyu Huang \\
  Tsinghua University\\
  Beijing, China \\
  \texttt{hsy17@mails.tsinghua.edu.cn} \\
   \And
   Bin Wang \\
   Huawei Noah's Ark Lab \\
   Beijing, China \\
   \texttt{wangbin158@huawei.com} \\
    \And
   Dong Li \\
   Huawei Noah's Ark Lab \\
   Beijing, China \\
   \texttt{lidong106@huawei.com} \\
   \AND
   Jianye Hao \\
   Huawei Noah's Ark Lab \\
   Beijing, China \\
   \texttt{haojianye@huawei.com} \\
   \And
   Ting Chen \\
   Tsinghua University \\
   Beijing, China \\
   \texttt{tingchen@tsinghua.edu.cn} \\
   \And
   Jun Zhu \\
   Tsinghua University \\
   Beijing, China \\
   \texttt{dcszj@tsinghua.edu.cn} \\
}

\begin{document}

\maketitle

\begin{abstract}
Circuit routing has been a historically challenging problem in designing electronic systems such as very large-scale integration (VLSI) and printed circuit boards (PCBs). The main challenge is that connecting a large number of electronic components under specific design rules involves a very large search space. 
Early solutions are typically designed with hard-coded heuristics, which suffer from problems of non-optimal solutions and lack of flexibility for new design needs. Although a few learning-based methods have been proposed recently, they are typically cumbersome and hard to extend to large-scale applications. In this work, we propose a new algorithm for circuit routing, named as Ranking Cost, which innovatively combines search-based methods (i.e., A* algorithm) and learning-based methods (i.e., Evolution Strategies) to form an efficient and trainable router. In our method, we introduce a new set of variables called cost maps, which can help the A* router to find out proper paths to achieve the global objective. We also train a ranking parameter, which can produce the ranking order and further improve the performance of our method. Our algorithm is trained in an end-to-end manner and does not use any artificial data or human demonstration. In the experiments, we compare with the sequential A* algorithm and a canonical reinforcement learning approach, and results show that our method outperforms these baselines with higher connectivity rates and better scalability.
\end{abstract}

\section{Introduction}

As described in Moore’s Law~\citep{schaller1997moore}, the number of transistors in a dense integrated circuit (IC) increases exponentially over time and the complexity of chips becomes higher and higher. Such high complexity makes the IC design a time-consuming and error-prone work. Thus more capable automatic design systems, such as electronic design automation (EDA) tools, are needed to improve the performance. In the flow of IC designs, we need to find proper paths to place wires which connect electronic components on ICs, and these wires need to achieve expected connectivity under certain constraints. One of the most important constraints is that wires on the same layout should not intersect. Besides, to reduce the signal propagation delay, the wire-length should be minimized. This is a critical and challenging stage in the IC design flow~\citep{hu2001survey}, known as circuit routing, which has been studied by lots of researchers~\citep{kramer1984complexity,he2020circuit}.

Circuit routing involves a large number of nets (a net is a set of vertices or pins with the same electrical property) to be routed, which is computationally expensive and makes manual design extremely time-consuming~\citep{coombs2001printed}. Even under the simplest setting, where only two pairs of pins need to be routed, it is an NP-complete problem~\citep{kramer1984complexity}. Although lots of circuit routing algorithms have been proposed ~\citep{zhang2016study}, there remain two major challenges: 
(1) Early solutions~\citep{hu2001survey} are typically designed with hard-coded heuristics, which suffer from problems of non-optimal solutions~\citep{zhang2016study} and lack of flexibility over new design needs. 
Therefore, a more powerful routing method that does not depend on domain knowledge is highly desired. 
(2) Although a few learning-based methods have been proposed~\citep{liao2020deep,he2020circuit} recently, their methods are hard to extend to large-scale applications. In real settings, there are lots of components and nets on a single chip, which shows greater demand for the scalability of routing algorithms. 

To relieve these problems, we propose a novel algorithm, denoted as Ranking Cost (RC), for circuit routing. We innovatively combine search-based methods (i.e., A* algorithm, the most commonly used algorithm in circuit routing tasks) and learning-based methods (i.e., Evolution Strategies~\citep{salimans2017evolution}, a powerful black-box optimization method) to form a trainable router with proper parametrization. 
Our method is flexible for integrating new constraints and rules so long as they can be merged into a global objective function. Moreover, 
our method is an one-stage algorithm, which optimizes the global objective function directly. In our method, we introduce a new set of variables called cost maps, which can help the A* routers to find out proper paths to achieve the global objective. We also train a ranking parameter, which can produce the ranking order and further improve the performance of our method. In the experiments, we compare our method with the commonly used A* method and a canonical reinforcement learning approach, and results show that our method outperforms these baselines with higher connectivity rates. Our ablation study also shows that the trained cost maps can capture the global information and lead to a reasonable routing solution. We also show that our method is scalable to larger maps. However, recent learning-based methods~\citep{liao2020deep,he2020circuit} only conduct experiments on small maps.


\begin{figure}[t]
\begin{center}
\subfloat[]{\begin{centering}
\includegraphics[width=0.3\linewidth]{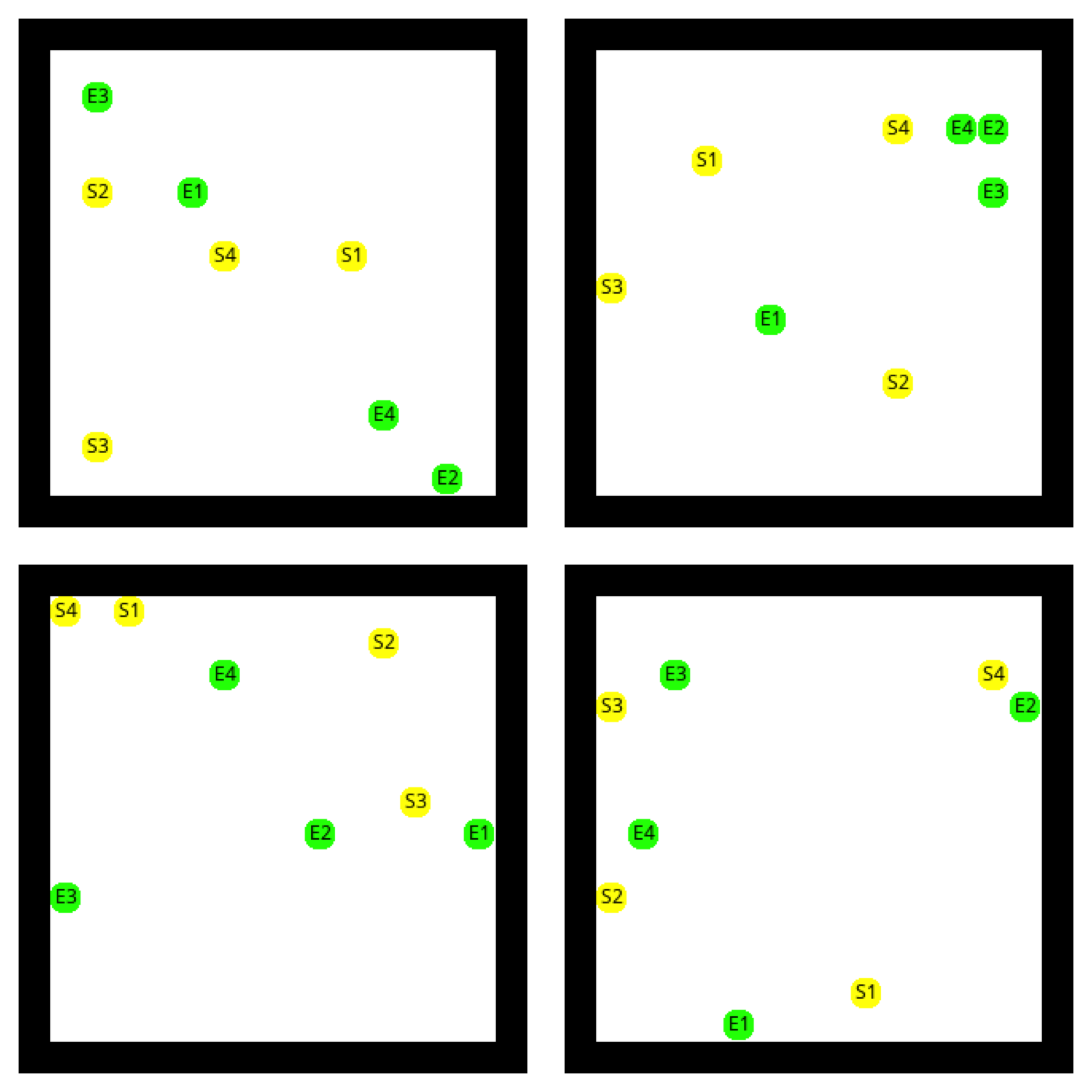}
\label{fig:start_s1}
\end{centering}
}\subfloat[]{\begin{centering}
\includegraphics[width=0.3\linewidth]{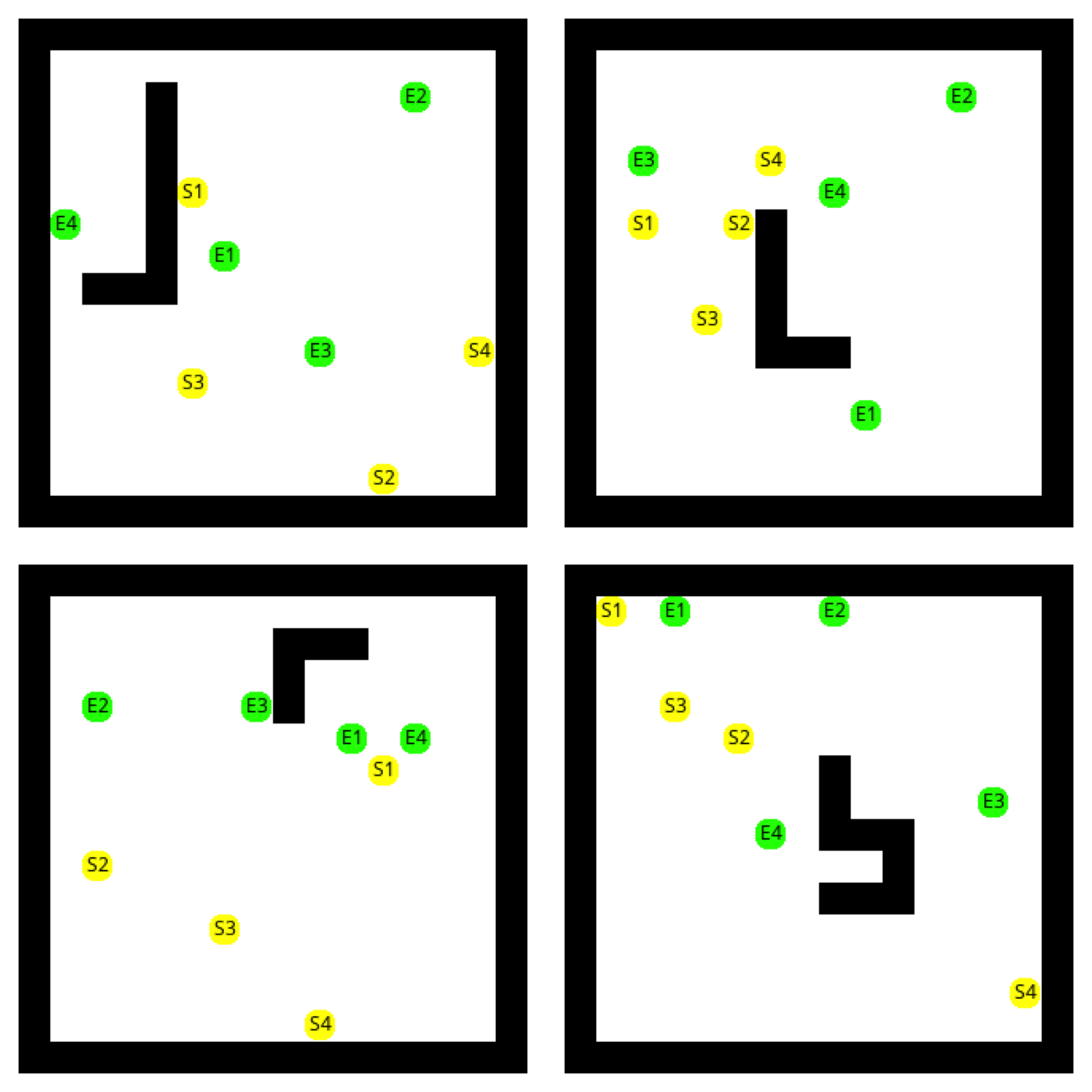}
\end{centering}
}\subfloat[]{\begin{centering}
\includegraphics[width=0.3\linewidth]{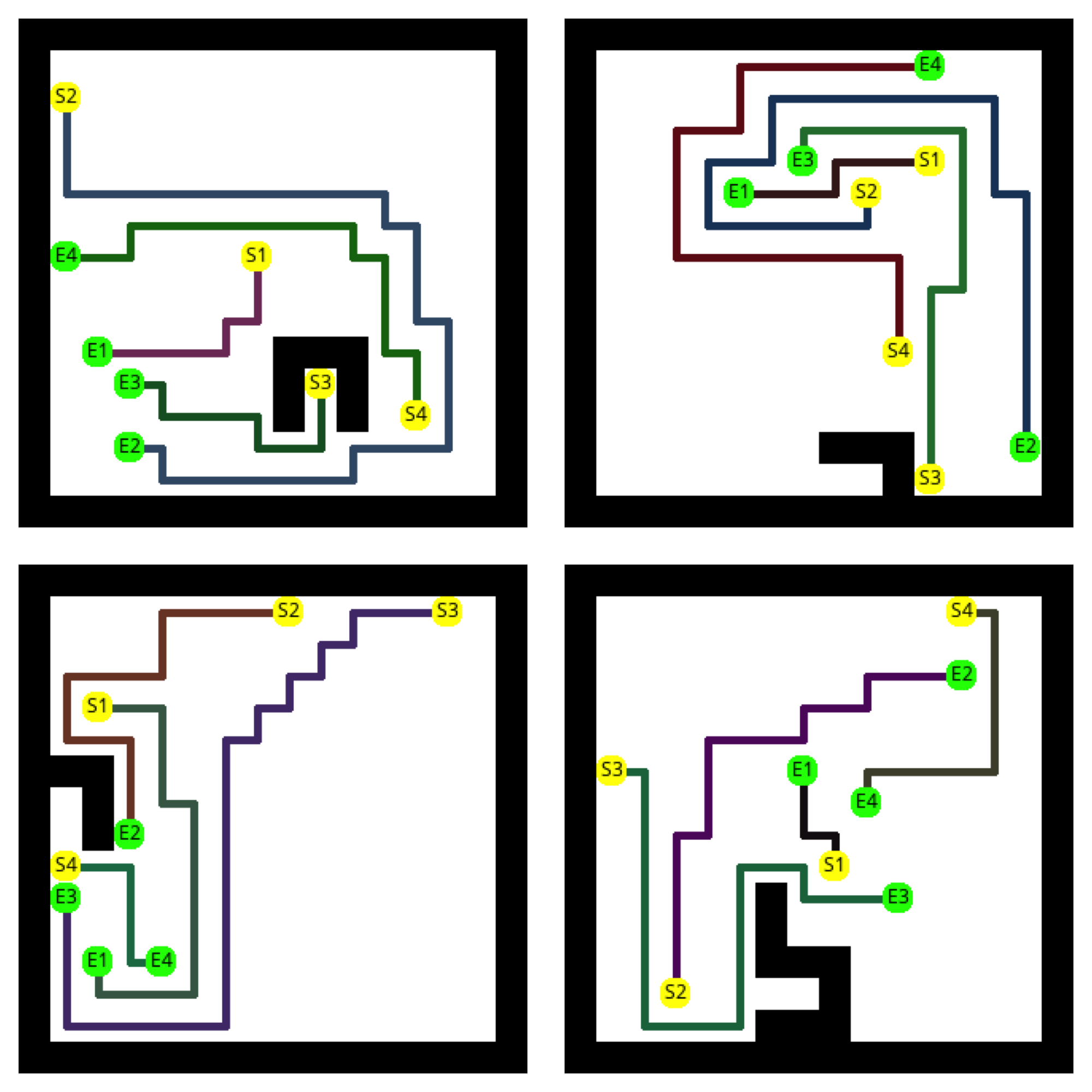}
\end{centering}
}

\subfloat[]{\begin{centering}
\includegraphics[width=0.3\linewidth]{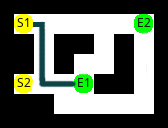}
\label{fig:start_s1}
\end{centering}
}\subfloat[]{\begin{centering}
\includegraphics[width=0.3\linewidth]{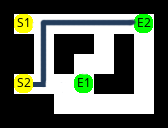}
\label{fig:start_s2}
\end{centering}
}\subfloat[]{\begin{centering}
\includegraphics[width=0.3\linewidth]{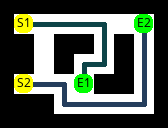}
\label{fig:solution_best}
\end{centering}
}
\end{center}
  \caption{Yellow vertices are the start vertices, green vertices are the end vertices and black blocks are obstacles where paths can not pass through. (a) Examples of maps without obstacle. (b) Examples of maps with obstacles. (c) Circuit routing results derived by our algorithm. 
In (d)-(f), $(S_i,E_i), i\in\{1,2\}$ are two pairs of vertices which need to be connected.
  (d) and (e) show that sequential A* algorithm will greedily find the shortest path of one pair, and it fails to obtain a global solution. (f) shows the solution found by our algorithm, which can connect the two pairs successfully.}
\label{fig:example}
\end{figure}
\section{Related Work}
\label{sec:related}
We summarize the related work on circuit routing.
\subsection{Heuristic Algorithms for Circuit Rouging}
\label{sec:seq}
The routing problem can be heuristically separated into two stages, the first being the global routing step~\citep{mo2001force,cho2007boxrouter2}, followed by detailed routing~\citep{ho1991general,mandal2020algorithms,chen2009global}. 
On the one hand, there are multiple heuristic-based approaches for global routing including regionwise routing~\citep{hu2001survey}, force-directed routing~\citep{mo2001force}, and rip-up and reroute~\citep{cho2007boxrouter2}. 
Besides, \citep{shi2016procedure} proposes a method to improve the distribution of congestion in global routing, and \citep{shi2017trapl} propose a framework to analyze local congestion for global routing. 
On the other hand, the most commonly used detailed routing algorithms are channel routing and its variants~\citep{ho1991general,mandal2020algorithms}, which decompose the routing region into routing channels and generate wires in these channels~\citep{chen2009global}. One main issue of the two-stage methods is that these two stages do not always coordinate well~\citep{shi2017trapl}, which results in enormous difficulty in joint optimization. Instead, our method is a one-stage algorithm and new design constraints can be simply involved in the objective function without changing the algorithm itself. 

A more straightforward strategy for circuit routing is to select a specific order and then route nets sequentially, e.g., sequential A* algorithm and Lees algorithm~\citep{huang2014ui,malavasi1993area}. The major advantage of this type of approaches is that the congestion information for previously routed nets can be taken into consideration while routing the current one. However, the drawback of these sequential approaches is that the quality of the solution is very sensitive to the orders~\citep{zhang2016study}. Moreover, earlier routed paths only focus on finding their own best solutions and are impossible to take into account the situation of subsequent paths. Such greedy strategies may make a solvable circuit routing problem insolvable. Figure~\ref{fig:example} shows an example that the sequential A* algorithm will fail to handle. In this example, there are two pairs of points to be routed, i.e., we should connect start vertices $S_i$ and end vertices $E_i, i\in\{1,2\}$, respectively. 
If we connect the pair $(S_1,E_1)$ first in its shortest path, it will make $(S_2,E_2)$ disconnected as shown in Figure~\ref{fig:example}(d), and vice versa (Figure~\ref{fig:example}(e)). But this case can be solved easily by our algorithm as shown in Figure~\ref{fig:example}(f), which means that our algorithm can take into account the global information when generating each path.

\subsection{Learning-based Methods for Circuit Routing}
There are some learning-based algorithms for circuit routing ~\citep{qi2014accurate,liao2020deep}. Early methods are more focusing on supervised learning, such as learning models to predict routing congestion~\citep{qi2014accurate} or predicting the routability~\citep{zhou2015accurate,xie2018routenet}. But the performances of these supervised models are constrained by the limits of human knowledge. Recently, ~\citep{liao2020deep} trained a deep Q-network (DQN) to solve the global routing problem and their method still suffers from the miscoupling problem of two-stage routing algorithms. Besides, \citep{he2020circuit} reform circuit routing to a tree search problem and utilize a deep neural network as the rollout policy. However, their rollout policy is trained in a supervised manner with artificial data and their search space is huge, which prevents it from scaling to larger applications. 
Instead, our method optimizes the global objective function directly and does not require any artificial data or human demonstration. Moreover, recent learning-based methods did not public their source codes and datasets. These papers lack key information and related materials, which prevents other researchers from re-implementing and following their work. Instead, we are the first learning-based work in this field with open source codes\footnote{Our codes can be found at \url{https://github.com/TARTRL/RankingCost}.}. For the EDA design community, our algorithm and source code can serve as a strong and fair baseline. For the AI community, our work provides a new evaluation benchmark for the combinatorial optimization problem.

\section{Background and Notations}
\label{sec:back}
In this section, we formalize the circuit routing problem first and then give a brief introduction to the OpenAI-ES algorithm~\citep{salimans2017evolution}.

\subsection{Circuit Routing}
Circuit routing is a path search problem, where the goal is to find non-intersecting paths that connect an arbitrary number of pairs of vertices. It can be formalized as a grid graph $G = (V, E)$, where each vertex $v_i \in V$ represents an intersection on the grid, and each edge $e_{ij}\in E$ represents the path between $v_i$ and its 1-hop neighbor $v_j \in V$. 
And the non-routable obstructive vertices form an obstacle set, denoted as $O\subset V$. 
In circuit routing, a net $N=\{v_{n1}, v_{n2},... \}\subset V$ is a set of vertices that need to be connected and a multi-vertex net can be decomposed into multiple two-vertex nets via a minimum spanning tree (MST) or a rectilinear Steiner tree (RST)~\citep{de1998rsr,hu2001survey}. Following \citep{he2020circuit}, we simplify this problem by letting each net only contain two vertices, i.e, a two-vertex net is defined as $N=\{v_s,v_e\}$ with a start vertex $v_s$ and an end vertex $v_e$. The net should be connected by a path $P = [v_1, v_2, ... , v_n]$ with $v_s = v_1$, $v_e = v_n$ and $P\cap O = \varnothing$. 
We use $|P|$ to represent the length of the path $P$. Given a set $\mathcal{N}$ of nets $\{N_i\}$, we need to find a set $\mathcal{P}$ of paths $\{P_i\}$ that connect these nets. A reasonable routing plan for a given set of nets is that all the paths do not share any vertex. In some cases, the routing problem does not have a solution because such a non-intersecting set $\mathcal{P}$ does not exist. In circuit routing, the commonly used objective is minimizing the total length of paths. Given a solution $\mathcal{P}$, the total length of paths is defined as $L=\sum_{P\in\mathcal{P}}|P|$. 

\subsection{Evolution Strategies}
\label{sec:es}
To optimize the parameters in our algorithm, we need a powerful black-box optimization method. Actually, evolution Strategies (ES) is a class of black-box optimization methods~\citep{rechenberg1994evolutionsstrategie,hansen2001completely}.
Recently, an evolution strategy variant, referred to as the OpenAI-ES~\citep{salimans2017evolution}, has attracted attention because it could rival the performance of modern deep reinforcement learning methods on multiple control tasks. In the OpenAI-ES, there is a reward function, denoted as $F(\psi)$, where $\psi$ is a solution vector, sampled from a probability distribution function $\psi\sim p_\theta(\psi)$. The goal is to maximize the expected value $J(\theta)$:
 \begin{eqnarray}
 \max_\theta J(\theta) = \mathbb{E}_{\psi\sim p_\theta}[F(\psi)].
 \end{eqnarray}

We can use the gradient ascent to optimize $\theta$:
 \begin{eqnarray}
 \theta \leftarrow \theta+\alpha\nabla_\theta J(\theta),
 \label{eq:ascent}
 \end{eqnarray}
 where $\alpha$ is the learning rate. In the OpenAI-ES, a score function estimator is used to calculate the gradient, which is similar to REINFORCE~\citep{williams1992simple}:
 \begin{eqnarray}
 \nabla_\theta J(\theta)=\nabla_\theta \mathbb{E}_{\psi\sim p_\theta}[F(\psi)] = \mathbb{E}_{\psi\sim p_\theta}[F(\psi)\nabla_\theta \log p_\theta(\psi)].
 \end{eqnarray}

Usually, $p_\theta$ is an isotropic multivariate Gaussian with mean $\theta$ and fixed covariance $\sigma^2I$. And the expected value can be rewritten as $\mathbb{E}_{\psi\sim p_\theta} [F(\psi)] = \mathbb{E}_{\epsilon\sim{\mathcal{N}(0,I)}}[F(\theta+\sigma\epsilon)]$. Thus the gradient estimator changes to:
 \begin{eqnarray}
 \nabla_\theta J(\theta)&=&\nabla_\theta \mathbb{E}_{\epsilon\sim \mathcal{N}(0,I)}[F(\theta+\sigma\epsilon)] \nonumber\\
 &=& \frac{1}{\sigma} \mathbb{E}_{\epsilon\sim {\mathcal{N}(0,I)}}[F(\theta+\sigma\epsilon)\epsilon],
 \label{eq:gradient}
 \end{eqnarray}
where $\epsilon$ is sampled from a standard normal distribution. Once we form an objective function $J(\theta)$, the gradient $\nabla_\theta J(\theta)$ can be approximated via Eq. ~(\ref{eq:gradient}) and parameters $\theta$ can be updated via Eq.~(\ref{eq:ascent}).

\section{Methodology}
\label{sec:method}
In this section, we first introduce the Ranking Learning algorithm and then the Cost-Map Learning algorithm. At last, we give out our final algorithm (Ranking Cost) and its training strategy. 

\subsection{Sequential A* with Ranking Learning}
\label{sec:rank}
The A* algorithm can be used to find the shortest path in a two-vertex net $N=\{v_s,v_e\}$, which is considered more efficient than breadth-first search. A* finds a path between a start vertex $v_s$ and an end vertex $v_e$, with a priority function defined as:
\begin{equation}
  s(v) = g(v)+h(v),
\label{eq:A_start}
\end{equation}
where $v$ is a vertex in the graph, $g(v)$ represents the total length from $v_s$ to the current vertex $v$, and $h(x)$ represents the future heuristic cost from $v$ to the end vertex $v_e$. The $h(x)$ is often set to the Euclidean distance or Manhattan distance. In a circuit routing problem, there are $k$ nets to be routed. Given a specific ranking order $R$, we can apply the A* algorithm to each net sequentially. For example, if there are 3 nets $\{N_1,N_2,N_3\}$ to be routed and a ranking order $R=(2,1,3)$ is given. We can apply A* to these 3 nets following the order $N_2\rightarrow N_1 \rightarrow N_3$. After the routing procedure, we can get a total path length from this ranking order, denoted as $L(R)$. Different ranking orders will result in different total lengths. We can go through all the ranking orders and get their corresponding total lengths. The ranking order with minimum total length can be used as the final solution. However, this is unacceptable in time complexity because a routing problem with $k$ nets has $O(k!)$ different ranking orders. 

To relieve the problem of combinatorial explosion of ranking orders, we propose a method to learn the ranking order. We define a $k$-dimension ranking parameter $\theta_r=\{\beta_1,\beta_2,...,\beta_k\}$. And the ranking order is determined completely by the ranking parameter. More concretely, the order of values in $\theta_r$ is exactly the routing order, and the net with the largest value will be routed first and the rest may be deduced by analogy. For example, given a ranking parameter $\theta_r=\{0.5,0.2,0.4\}$, we have $\beta_1>\beta_3>\beta_2$ and the corresponding ranking order will be $R_{\theta_r}=(1,3,2)$. And the ranking parameter will determine the final routing result and also the total path length. We define a reward function over the ranking parameter:
\begin{equation}
  F(\theta_r) = -L(R_{\theta_r}),
\label{eq:f_rank}
\end{equation}
where $L(R_{\theta_r})$ is the total length when taking the order $R_{\theta_r}$, and less length leads to larger reward. As described in Section~\ref{sec:es}, we define the expected value over $F(\cdot)$ as:
\begin{equation}
J(\theta_r) = \mathbb{E}_{\psi\sim p_{\theta_r}}F(\psi)=\mathbb{E}_{\epsilon\sim {\mathcal{N}(0,I)}} F(\theta_r+\sigma\epsilon).
\end{equation}
The gradient $\nabla_{\theta_r}J(\theta_r)$ can be estimated via Eq.~(\ref{eq:gradient}), and $\theta_r$ can be updated via Eq.~(\ref{eq:ascent}). 
At every training step, we sample a fixed number of noise vectors from Gaussian, add them to the ranking parameter and then evaluate all the new ranking parameters. The reward function $F(\cdot)$ will return a group of scalar rewards and the ranking parameter will be updated with these rewards. In this way, we can train a ranking parameter $\theta_r$ to approach the minimal total length $L$.

As mentioned in Section~\ref{sec:seq}, a sequential routing algorithm may make a solvable circuit routing problem insolvable. And even a method with a learned ranking order will still suffer from this problem. In the following section, we will introduce how our method could overcome this problem by coupling the A* algorithm with cost maps.

\subsection{Circuit Routing with Cost Maps}
As introduced in the previous section, $g(v)$ in Eq.~(\ref{eq:A_start}) represents the total length from start vertex $v_s$ to current vertex $v$, thus the A* algorithm can only search for the shortest path of the current net and it cannot take into account the information of following paths. Our goal is to add global information to the A* routers so that these routers can cooperate to achieve their common objective. First, for each net $N_i, i\in\{1,...,k\}$, we define its cost map as $C_i=\{c_1,c_2,...,c_m\}\in \mathbb{R}^m$, where $m=|V|$ is the number of vertices in graph $G$. Therefore, there is a total of $k$ cost maps, and all the cost maps can be learned with our algorithm. We apply a cost-map function $C(i,v)=c,c\in C_i$ to simplify the notation. In this way, we can reformulate the A* algorithm by adding the cost maps: 
\begin{equation}
  s(v)=g(v)+\sum_{i=1}^{k}C(i,v)+h(v),
\label{eq:A_cost}
\end{equation}
where $g(v)$ and $h(v)$ are with the same definitions as in Eq.~(\ref{eq:A_start}). The cost-map enhanced A* algorithm reveals two key points: (1) When removing $g(v)$ and $h(v)$ from Eq.~(\ref{eq:A_cost}), it will degenerate into a pure learning-based algorithm and the path will be determined completely by the cost function.
(2) When removing the cost functions or set them to zero (i.e., $C(i,v)=0$), it is a pure search-based algorithm. Hence, it is a method which combines the searched-based algorithm and the learning-based algorithm and could take advantage of both sides. Firstly, the A* search makes our method easier to find a connected path in a complex environment, while it is hard for learning-based methods to find a connected path~\citep{tamar2016value}. Second, the global information can be merged into cost maps and the global objective can be approached by tuning the cost maps. 

To learn the cost maps, we first define a cost-map parameter $\theta_c\in \mathbb{R}^{k\times m}$ and the values in cost maps are defined as $c_j = \max(0,\theta_c[i,j]),c_j \in C_i$. Given a ranking order $R$ and cost maps, we can route the nets sequentially using the cost-map enhanced A* and a total path length $L$ will be obtained from the routing result. Similar to Eq.~(\ref{eq:f_rank}), we define a reward function over the cost-map parameter and also its the expected value:
\begin{equation}
\begin{split}
F(\theta_c) &= -L(R,C_{\theta_c}), \\
J(\theta_c) &= \mathbb{E}_{\psi\sim p_{\theta_c}}F(\psi) = \mathbb{E}_{\epsilon\sim N(0,I)}F(\theta_c+\sigma\epsilon),
\end{split}
\end{equation}
where $C_{\theta_c}$ presents the cost maps derived from $\theta_c$, and $\theta_c$ can be updated via Eq.~(\ref{eq:ascent}). By iteratively executing the A* search and cost-map learning step, our method could solve some hard problems that a sequential routing algorithm can not solve as shown in Figure~\ref{fig:example}. In the next section, we will introduce how cost-map learning can couple with ranking learning and how to train the algorithm efficiently in practice.

\begin{algorithm}[t]
\caption{Ranking Cost Training}
\label{algo:RC_training}

{\bf Initialize:} Graph $G=(V,E)$, vertex number $m=|V|$, nets $\{N_1,...,N_k\}$, ranking parameter $\theta_r\in \mathbb{R}^k$, cost-map parameter $\theta_c\in \mathbb{R}^{k\times m}$, learning rate, $\alpha$, noise standard deviation $\sigma_r$ and $\sigma_c$, evaluator number (population size) $n$.\\
\For{$t=1,2,...$}
{
  \For{$i=1,...,n$}
  {
    Sample $\epsilon_i\sim N(0,I)$.\\
    Get sampled ranking parameter $\hat{\theta}_r$ and cost maps parameter $\hat{\theta}_c$ with:
      \begin{equation}\nonumber
        \begin{split}
          \hat{\theta}_r &= \theta_r+\sigma_r\epsilon_i,\\
          \hat{\theta}_c &= \theta_c+\sigma_c\epsilon_i
      \end{split}
      \end{equation}\\
    Construct ranking order $R$ and cost maps $C$ from $\hat{\theta}_r$ and $\hat{\theta}_c$.\\
    Route nets on graph $G$ with the ranking order and cost maps.\\
    Compute rewards $r_i = F(\hat{\theta}_r,\hat{\theta}_c)$ from routing results.
  }
  Collect all scalar reward $\{r_1,...,r_n\}$ from each evaluator.\\
  Compute normalized rewards $\{\overline{r}_1,...,\overline{r}_n\}$ via Eq.~(\ref{eq:r_normolize}).\\
  Update ranking parameter $\theta_r$ and cost map parameter $\theta_c$ with:
    \begin{equation}\nonumber
    \begin{split}
      \theta_r &\leftarrow \theta_r+\alpha\frac{1}{n\sigma_r}\sum_{j=1}^n \overline{r}_j\epsilon_j\\
      \theta_c &\leftarrow \theta_c+\alpha\frac{1}{n\sigma_c}\sum_{j=1}^n \overline{r}_j\epsilon_j
    \end{split}
    \end{equation}
}
\end{algorithm}

\subsection{Ranking Cost Algorithm}
Our final algorithm, denoted as Ranking Cost (RC), can learn the ranking parameter $\theta_r$ and cost-map parameter $\theta_c$ jointly. In Eq.~(\ref{eq:A_cost}), all the cost maps are used when calculating the priority function and the ranking order has no impact on the cost maps. To learn the cost-map with ranking order, we change the priority function to:
\begin{equation}
  s_j(v) = g_j(v)+\sum_{i=j+1}^k C(i,v)+h_j(v),
\end{equation}
where $s_j(v)$ is the priority function for the net whose ranking order is the $j$-th. This modification is reasonable because the earlier routed nets will use more cost maps and observe more global information. For example, when routing the first net, its ranking index is $j=1$ and it will use all the rest $k-1$ cost maps. As a result, the first routed path focuses more on its impact on other unrouted nets. When routing the last net, its ranking index is $j=k$ and no cost map will be used. As a result, the last net is routed via a normal A* algorithm and the searched solution is the shortest path at the current situation. As all the previous nets have been routed, the best way to route the last net is just to find out the shortest path. Our RC algorithm achieves this naturally and makes it flexible for arbitrary ranking orders. When the ranking order changes, the order and the usage of cost maps change accordingly. In this way, the reward function can be defined as:
\begin{equation}
  F(\theta_r,\theta_c)=-L(R_{\theta_r},C_{\theta_c}),
\end{equation}
and both $\theta_r$ and $\theta_c$ can be updated via the OpenAI-ES algorithm. Actually, the reward function $F$ can be integrated with arbitrary metrics, such as the signal latency. Thus the cost maps are free to adapt to new constraints and design rules. 

In the OpenAI-ES algorithm, we will sample a fixed number of Gaussian noises and add them to the original parameters to form new parameters. These new parameters are then fed into $n$ evaluators (or workers). The evaluator has an independent environment and executes the algorithm based on received parameters. Finally, each evaluator will return a scalar reward. To stabilize the training process, we normalize collected rewards $\{r_1,...,r_n\}$ from all the evaluators:
\begin{equation}
  \overline{r}_i = (r_i-r_{mean})/r_{std},
\label{eq:r_normolize}
\end{equation}
where $r_{mean}$ and $r_{std}$ are the mean and standard deviation of all the rewards.

To further improve the final results, we add a post-processing step to the solution obtained from the Ranking Cost algorithm. In the post-processing step, we will choose a connected path and re-plan it using the canonical A* algorithm while keeping other paths fixed. If we find a shorter path during the re-planning, we will update the current path to the new one. This post-processing step will be iteratively applied on each path until all the paths can not be updated. Algorithm~\ref{algo:RC_training} shows the overall training procedure of our RC algorithm. In practice, all the evaluators can execute in parallel as proposed in ~\citep{salimans2017evolution}.

\section{Experiments}
\label{sec:exp}
In this section, we will compare our proposed method with some other baselines. We will also study how ranking learning and cost maps work in our algorithm. Finally, we further show the scalability of our algorithm for larger applications.

\begin{table*}[t]
\begin{center}
    \begin{tabular}{c|c|c|c}
      \hline
    & $16\times 16$ (4 pairs) & $32\times 32$ (32 pairs)& $64\times 64$ (10 pairs) \\
    \hline
    & no obstacle & no obstacle & no obstacle \\
    \hline
    Seq A*(5) &0.92±0.02(37.0$\pm$0.16)&  0.96±0.01(109.9$\pm$0.36)&  0.76±0.04(434.2$\pm$5.06)\\
    Seq A*(200)& 0.96±0.0(36.5$\pm$0.0)&  1.0±0.0({\bf 107.0}$\pm$0.0)&  0.84±0.0({\bf 400.0}$\pm$1.11)\\
VINs& 0.86±0.0(37.1$\pm$0.27)&  0.52±0.03(124.1$\pm$1.3)&  0(-)\\
Cost Learning& 0.93±0.0(38.1$\pm$0.02)& 0.98±0.0(111.3$\pm$2.0)& 0.88±0.0(457.7$\pm$0.0)\\
Ranking Cost I& 1.0±0.0(36.5$\pm$0.02)&  1.0±0.0(108.2$\pm$0.32)&  0.86±0.0(417.1$\pm$0.0)\\
Ranking Cost II & {\bf 1.0}±0.0({\bf 36.5}$\pm$0.02)&  {\bf 1.0}±0.0(108.2$\pm$0.34)&  {\bf 0.86}±0.0(413.8$\pm$0.0)\\
      \hline
          & with obstacles & with obstacle & with obstacle \\
    \hline
Seq A*(5)& 0.92±0.01(36.3$\pm$0.09)& 0.53±0.02(105.4$\pm$0.73)& 0.1±0.03(384.0$\pm$13.65)\\
Seq A*(200)& 0.96±0.0(36.1$\pm$0.0)& 0.78±0.0(103.8$\pm$0.0)& 0.26±0.01(367.5$\pm$0.0) \\
VINs& 0.87±0.01(36.4$\pm$0.16)& 0.21±0.02(114.7$\pm$2.09)& 0(-)\\
Cost Learning& 0.9±0.0(36.8$\pm$0.09)& 0.67±0.01(104.4$\pm$0.31)& 0.16±0.0(405.5$\pm$0.0)\\
Ranking Cost I&  0.98±0.0(36.1$\pm$0.02)&  0.82±0.0(103.8$\pm$0.0)& 0.32±0.0(360.5$\pm$0.0)\\
Ranking Cost II& {\bf 0.98}±0.0({\bf 36.1}$\pm$0.02)& {\bf 0.82}±0.0({\bf 103.8}$\pm$0.0)& {\bf 0.32}±0.0({\bf 358.5}$\pm$0.0)\\
\hline
    \end{tabular}
  \end{center}
  \caption{Evaluation results of different algorithms over 6 seeds. Two values and their corresponding standard deviations are reported. The first value is the success rate (the higher the better) and the second value in the bracket is the common average length (the lower the better). Our methods achieve the best performance on most of the maps. The comparison between CML and RC shows that the ranking parameter learning can improve the performance of our algorithm.}
  \label{table:compares_std}
\end{table*}

\begin{table*}[t]
\begin{center}
    \begin{tabular}{|c|c|c|c|c|c|}
      \hline
    Seq A*(5) &  Seq A*(200) & VINs & Cost Learning& Ranking Cost I & Ranking Cost II\\
    \hline
    0.8 sec/map & 41.5 sec/map& 1.4 sec/map& 37.5 sec/map& 38.3 sec/map& 38.3 sec/map\\
    \hline
    \end{tabular}
  \end{center}
  \caption{The wallclock time of different algorithms. The Ranking Cost algorithm takes more running time,
   since it involves a learning procedure when solving each task. But the time consumption is acceptable, i.e., our algorithm takes less than one minute to get the solution,  
  and it is worthy to sacrifice more time to earn better solutions.}
  \label{table:time}
\end{table*}

\subsection{Methods Evaluation}
\label{sec:set_up}
To evaluate our algorithm, we build a grid map simulator as the test environment. We construct 300 maps with three different sizes: $16\times 16$, $32\times 32$ and $64\times 64$. These maps are split into two types, i.e., a simple one without obstacle and a more complex one with obstacles. Figure~\ref{fig:example}(a) and Figure~\ref{fig:example}(b) show some of the maps used in our experiments. Algorithms will be evaluated on these maps, and the success rates and the average of total lengths will be reported. Our experiments are performed on a desktop machine with 128 GB RAM and one 64-core CPU. More details about the maps can be found in the supplementary material.

\noindent{\bf Baselines}:\\
{\bf Sequential A*}: Sequential A* is a common used search-based algorithm in circuit routing. It routes nets with a specific ranking order and different orders will lead to different results. In the experiments, we randomly sample 5 different ranking orders and run the algorithm 5 times. The best score of 5 runs will be reported. To build a stronger baseline, we further randomly sample 200 different ranking orders and the best score of 200 runs will be reported.\\
{\bf Value Iteration Networks (VINs)}~\citep{tamar2016value}: The VIN is a learning-based approach for routing on grid maps. VINs use the value iteration from reinforcement learning and can achieve better performance compared with supervised models. 
We apply VINs to the circuit routing problem by sequentially executing it on each net. Similar to the sequential A* algorithm, we randomly sample five different ranking orders and the best score of five runs will be reported.\\
{\bf Cost-Map Learning (CML)}: We construct a Cost-Map Learning algorithm from the Ranking Cost. In the CML, the ranking parameter will not be trained (the ranking order is fixed) and only the cost-map will be trained.\\

The Ranking Cost as presented in Algorithm~\ref{algo:RC_training} is our final method for circuit routing problem and more details about its hyper-parameters can be found in the supplementary material. We also implement two versions of the Ranking Algorithm, i.e., the version without post-processing step is denoted as Ranking Cost I and the version with post-processing step is denoted as Ranking Cost II. Table~\ref{table:compares_std} shows the evaluation results of different algorithms on all the maps over 6 seeds. Two values and their corresponding standard deviations are reported for each algorithm. The first value is the success rate (the higher the better) and the second value in the bracket is the common average length (the lower the better). Our final algorithm (Ranking Cost) achieves the best performance on most of the maps. On the most complex maps (with size $64\times 64$, obstacles and 10 pairs of vertices), Ranking Cost outperforms other baselines by a large margin, which implies that our method has greater advantage on handling larger maps. The comparison between Seq A*$_{(200)}$ and Ranking Cost shows that our method can achieve the performance which the fully sampled A* will never achieve. Taking maps with the size of 16x16 (4 pairs) for example, there are only $4!=24$ orders, and we sampled all the possible orders for the A* since our maximum sampling number for is 200. But the final connecting rate of the A* is $96\%$. However, our RC algorithm can reach $100\%$ connecting rate. Even given the full sampling, the A* will fail in many cases. Moreover, the comparison between Cost-Map Learning and Ranking Cost shows that the ranking parameter learning can further improve the performance of our algorithm. The comparison between Ranking Cost I and Ranking Cost II shows that our post-processing step can help to reduce the total connected length with insignificant time-consuming.

\begin{figure*}[t]
\begin{center}

\subfloat[]{\begin{centering}
\includegraphics[width=0.23\linewidth]{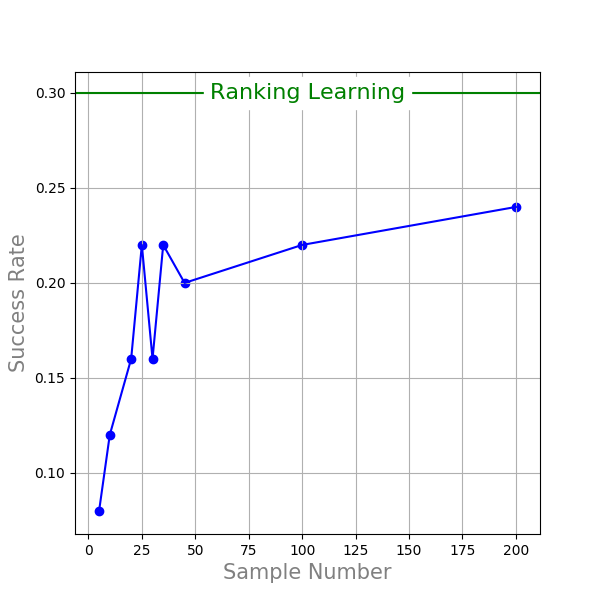}
\label{fig:start_s1}
\end{centering}
}\subfloat[]{\begin{centering}
\includegraphics[width=0.23\linewidth]{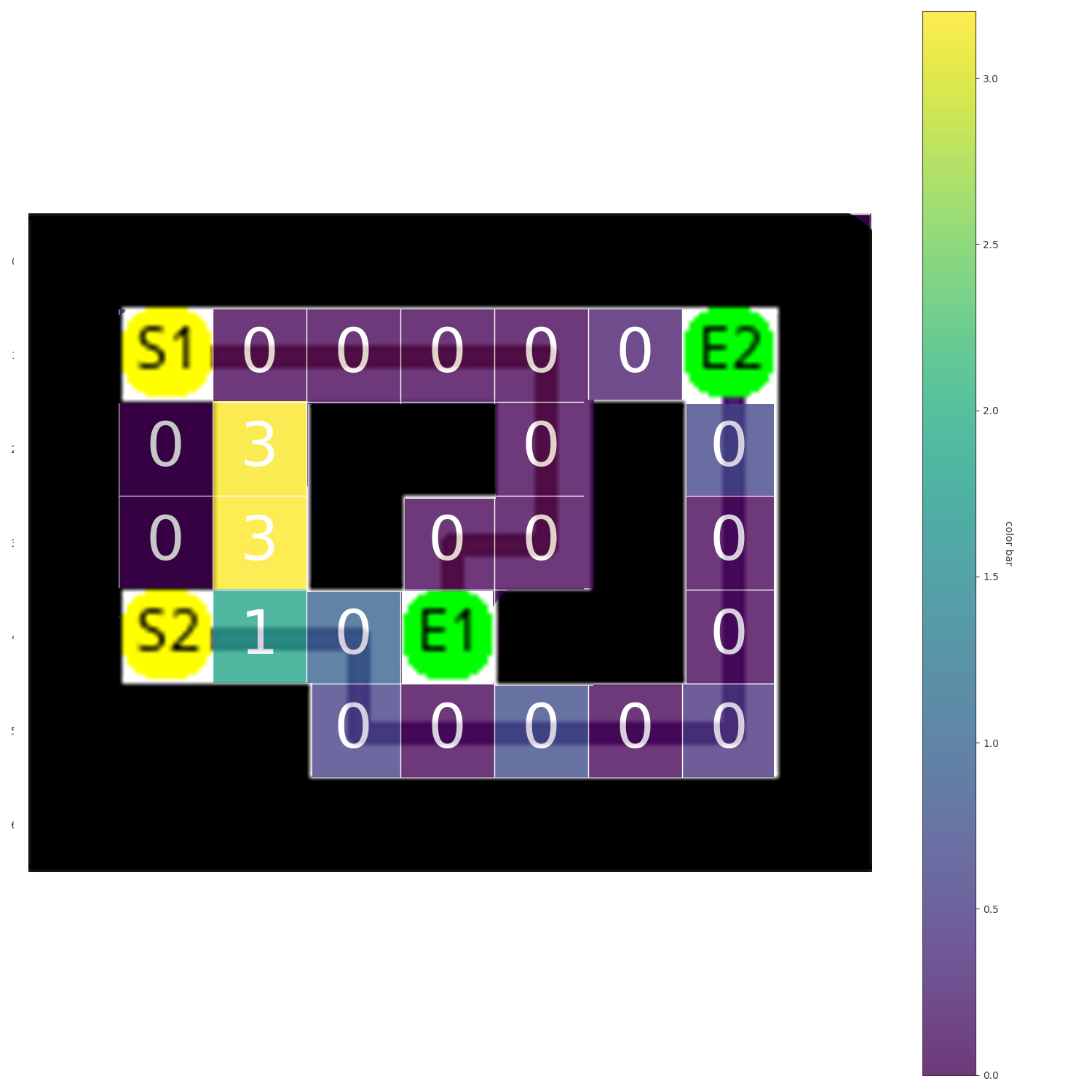}
\label{fig:solution_best}
\end{centering}
}\subfloat[]{\begin{centering}
\includegraphics[width=0.23\linewidth]{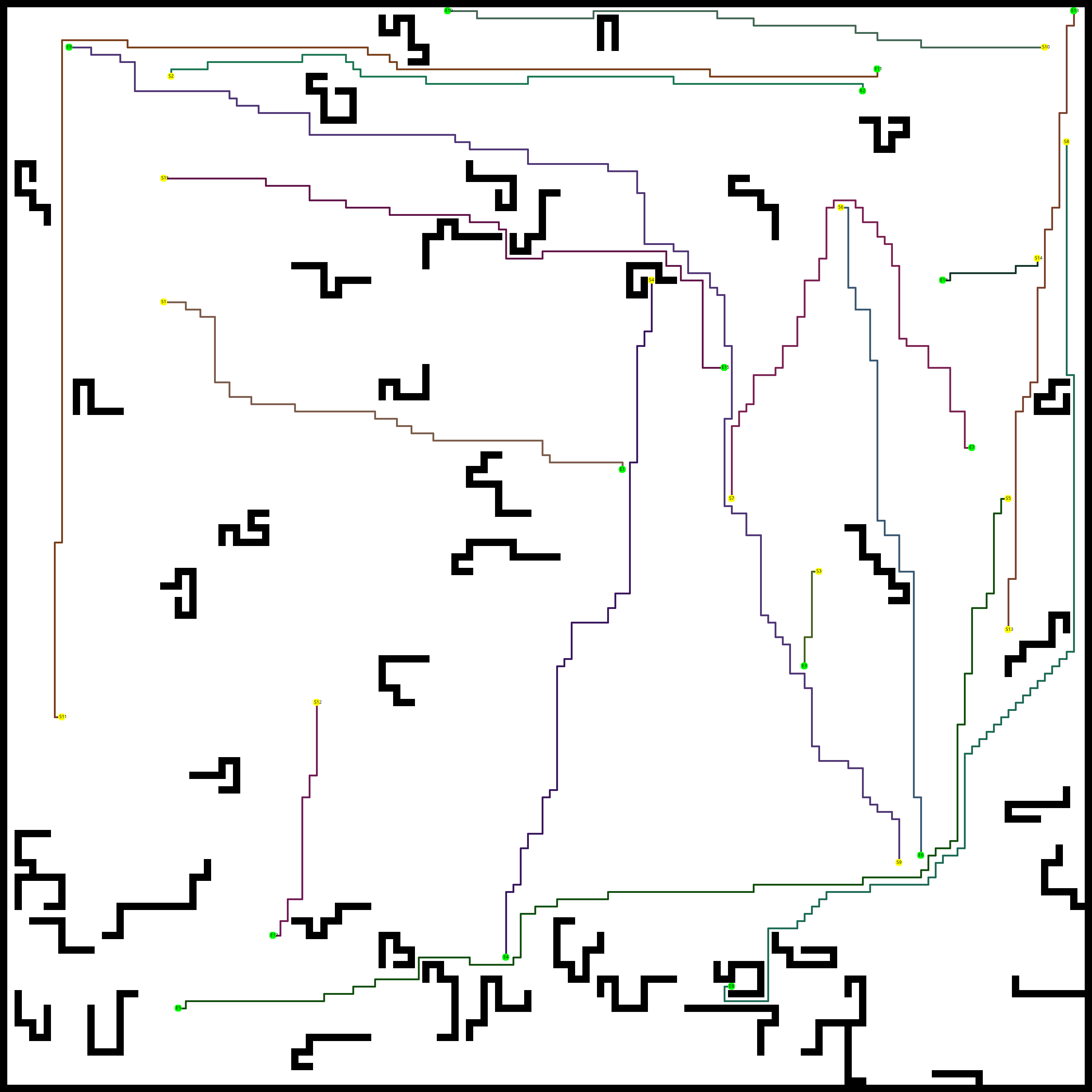}
\label{fig:big1}
\end{centering}
}\subfloat[]{\begin{centering}
\includegraphics[width=0.23\linewidth]{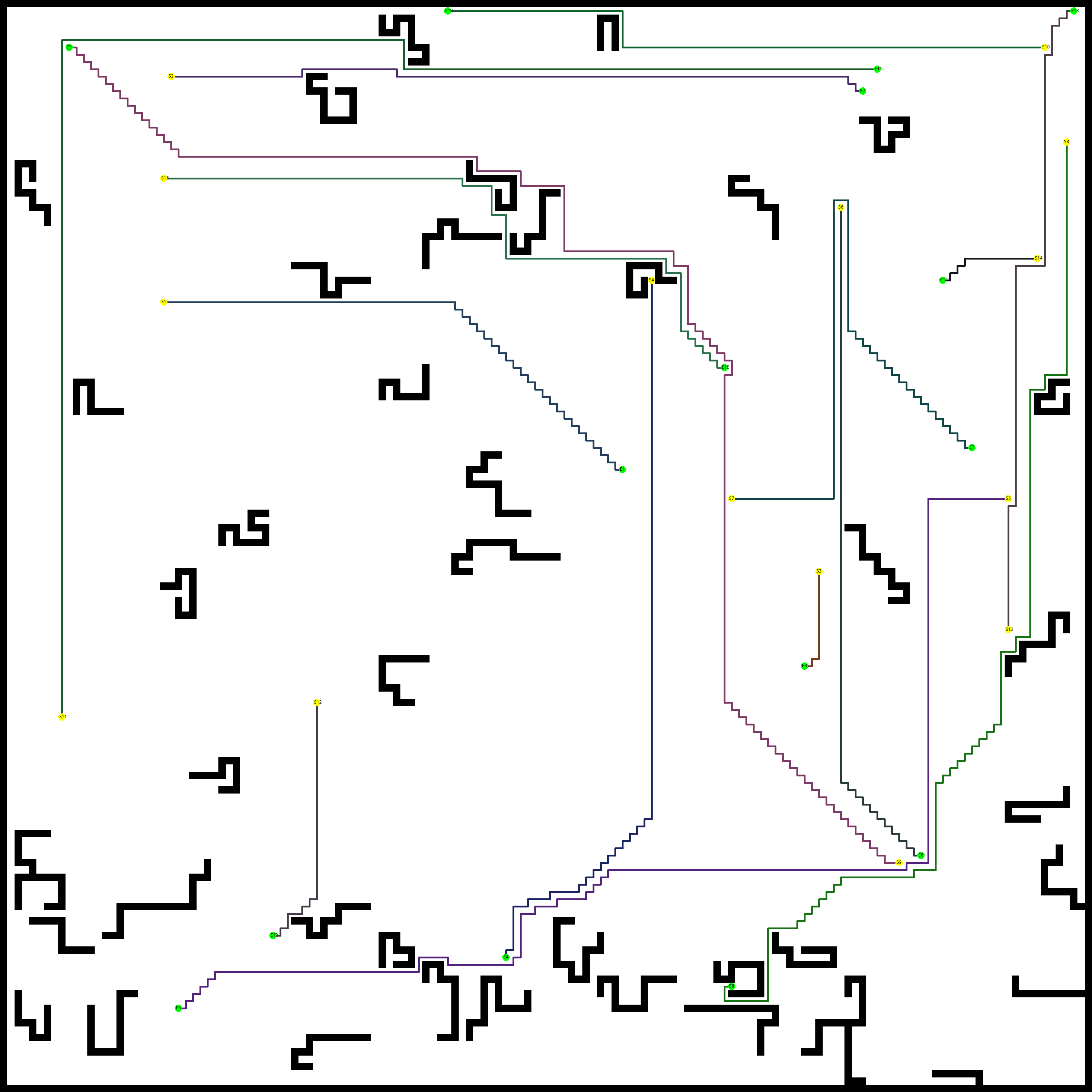}
\label{fig:big2}
\end{centering}
}
\end{center}
 \caption{(a) The curve of success rates and sample numbers and the experiment is conducted on maps with size of $64\times64$ and with obstacles. Our learned ranking order achieves better performance than randomly sampled orders. (b) Trained cost map of the example from Figure~\ref{fig:example}(f). There are two larger cost values (the yellow grids) in the cost map, which will give out a large cost if the path chooses to go through here. As a result, the first pair ($(S_1,E_1)$) will not take its shortest path which will block the second pair($(S_2,E_2)$). It shows that cost maps can capture global information and guide routers to complete the task successfully. (c) An example of routing results of the Ranking Cost algorithm on a large map with the size of $150\times 150$ and with 15 pairs, which implies that our method could be adapted to larger scales of applications. (d) An example shows the routing result of the Ranking Cost algorithm with the post-processing step, which implies that the post-processing step could improve the final planning result.}
\label{fig:exp}
\end{figure*}
\subsection{Ablation Study}

In this section, we study how each part of our algorithm works and the scalability of our algorithm.

\subsubsection{Impact of Ranking Orders}
To show the impact of ranking learning, we increase the sample number of orders for sequential A* and use the Ranking Learning algorithm described in Section~\ref{sec:rank} as a comparison. Figure~\ref{fig:exp}(a) shows the curve of the changes in success rates with different order sample numbers. The result shows that we can improve the success rate by increasing the order sample number, but it is hard to further improve the performance by linearly increasing the sample number as the order complexity is $O(k!)$, where $k$ is the number of pairs (or nets). However, our learned ranking order can achieve much better performance than randomly sampled orders.

\subsubsection{Impact of Cost Maps}
To study the impact of cost maps, we fix the ranking order and only train the cost-map parameter. We use the example map showed in Figure~\ref{fig:example}(f) and the ranking order is fixed as $R=(1,2)$. Because the ranking order is fixed and there are only two cost maps, only the second cost map will be used when routing the first pair (i.e., $(S_1,E_1)$). Finally, we visualize the trained cost maps as shown in Figure~\ref{fig:exp}(b). There are two larger cost values (the yellow grids) in the cost map, which will give out a large cost if the path chooses to go through here. As a result, the first pair ($(S_1,E_1)$) will take the path as shown in Figure~\ref{fig:example}(f) instead of taking its shortest path which will block the second pair($(S_2,E_2)$). It shows that our cost maps can capture global information and guide routers to approach global optima.

\subsubsection{Scalability of Ranking Cost Algorithm}
In \citep{liao2020deep}, their algorithm is only evaluated on the maps with the size of $8\times 8$. In \citep{he2020circuit}, their algorithm is evaluated on the maps with the size of $30\times 30$ and with only 5 pairs. The search space is huge in their methods, which prevents them from applying to larger maps. To test the scalability of our algorithm, we apply our algorithm on larger maps with the size of $150\times 150$ and with 15 pairs. Our algorithm solves such large maps within 3 min and Figure~\ref{fig:exp}(c) shows the routing result of our method without post-processing step, which implies that our method could be adapted to larger scales of applications. Figure~\ref{fig:exp}(d) shows the routing result of the Ranking Cost algorithm with the post-processing step, which implies that the post-processing step could improve the final planning result. 

\section{Discussion}
\label{sec:dis}
In this paper, we propose a novel algorithm, called Ranking Cost, to solve the historically challenging circuit routing problem. In our method, we innovatively combine search-based algorithms and learning-based algorithms to form an efficient and trainable router under a proper parameterization. Our method is a one-stage circuit routing algorithm which can optimize the global objective function directly, and it is easy to implement and flexible for new design rules and constraints. Experimental results show that our method is powerful and scalable to more complex tasks. In the future, we will try to explore faster optimizer for the Ranking Cost parameters and extend our algorithm to broader applications, such as pedestrian path prediction and robot navigation. Lastly, we do not believe our work has broader negative societal impacts, as we focus on developing circuit routing algorithms with evolution-based methods.


\bibliography{nips19.bib}
\bibliographystyle{nips}

\appendix

\section{Environment Setup}
\label{ap:env}
To evaluate our algorithm, we build a grid map simulator as the test environment. Algorithms are evaluated on randomly generated grid maps. In the experiment, we take use of three different maps sizes: $16\times 16$, $32\times 32$ and $64\times 64$. We construct two types of maps, i.e., a simple one without obstacles and a more complex one with obstacles. All the start vertices and end vertices are also randomly generated in both simple maps and complex maps. Each map with size of $16\times 16$ contain 4 start-end pairs, and 6 pairs for $32\times 32$ and 10 pairs for $64\times 64$. We randomly generate 50 maps for each size and each type and there are total 300 random maps. We report the success rate and the common average of total length of each algorithm. The success rate is the proportion of successfully connected maps to all random maps. Because different algorithm will have different connected maps, it is not fair to compare the average length over each algorithm's own successful maps. A algorithm with only one successful case may have lower average length than a algorithm with $100\%$ success rate. Thus we only report the average length over common successful maps of all algorithms.  All the maps can be found in the supplementary material.

\section{Details of Hyper-parameters}
\label{ap:alg_detail}

 For both Cost-Map Learning and Ranking Cost, the maximum training episode is $1000$, learning rate is $0.001$, the number of evaluators is 40, the noise standard deviation $\sigma_r$ is 0.1 and $\sigma_c$ is 0.1. All the rewards returned from evaluators will be scaled to $[-1,0]$. When there is not a successful connection, a reward of $-1$ will be given. Our code can be found in the supplementary material.
 
\section{Result of Ranking Orders}
 \label{ap:ranking_order}
\begin{table}[h]
\renewcommand\tabcolsep{3.0pt}   
\begin{center}
        \begin{tabular}{|c|cccccccccccc|c|c}
            \hline
            Sample Number & 5   & 10   &15  &20  &25  &30  &35  & 40  & 45 & 50 & 100 & 200 & Ranking Learning\\
            \hline
            Success Rate &  0.08& 0.12 &0.16&0.16&0.22&0.16&0.22& 0.20& 0.20& 0.22 & 0.22 & 0.24 & 0.30 \\  
            \hline
        \end{tabular}
    \end{center}
    \caption{The changes of success rates with different order sample numbers. The result shows that it is hard to improve the performance by linearly increasing the sample number. The learned ranking order can achieve much better performance than randomly sampled orders.}
    \label{table:samples}
\end{table}

 \section{Other Routing Tasks}
There are also some other routing problems, such as routing of hydraulic systems~\citep{chambon1991automated}, routing of ship pipes~\citep{kang1999design}, routing of urban water systems~\citep{christodoulou2010pipe,grayman1988modeling}, and routing of city logistics~\citep{barcelo2007vehicle,ehmke2012advanced}. Multi-agent path finding, a task similar to circuit routing, has been exhaustively studied in robotics and video games~\citep{silver2005cooperative,sturtevant2014grid,babayan2018belief,Atzmon2020Multi}. But in a multi-agent path finding task, paths are allowed to intersect as long as the agents do not appear in the same place at the same time. To a certain degree, circuit routing is more difficult than multi-agent path finding tasks.


\end{document}